# Skin Cancer Segmentation and Classification with NABLA-N and Inception Recurrent Residual Convolutional Networks

**Md Zahangir Alom, Theus Aspiras, Tarek M. Taha, and Vijayan K. Asari**
Department of Electrical and Computer Engineering, University of Dayton, OH 45469, USA.
e-mail: {alomm1, aspirast1, ttaha1, vasari1}@udayton.edu

## Abstract

*In the last few years, Deep Learning (DL) has been showing superior performance in different modalities of biomedical image analysis. Several DL architectures have been proposed for classification, segmentation, and detection tasks in medical imaging and computational pathology. In this paper, we propose a new DL architecture, the NABLA-N network ($\nabla^N$-Net), with better feature fusion techniques in decoding units for dermoscopic image segmentation tasks. The $\nabla^N$-Net has several advances for segmentation tasks. First, this model ensures better feature representation for semantic segmentation with a combination of low to high-level feature maps. Second, this network shows better quantitative and qualitative results with the same or fewer network parameters compared to other methods. In addition, the Inception Recurrent Residual Convolutional Neural Network (IRRCNN) model is used for skin cancer classification. The proposed $\nabla^N$-Net network and IRRCNN models are evaluated for skin cancer segmentation and classification on the benchmark datasets from the International Skin Imaging Collaboration 2018 (ISIC-2018). The experimental results show superior performance on segmentation tasks compared to the Recurrent Residual U-Net (R2U-Net). The classification model shows around 87% testing accuracy for dermoscopic skin cancer classification on ISIC2018 dataset.*

## 1. INTRODUCTION

Cancer is one of the main causes of death all around the world. Out of the many types of cancer, melanoma is a type of cancer which affects the skin (mostly in pigment cells). It is reported that 1 out of 33 men and 1 out of 52 women are affected by melanoma skin cancer in the USA alone and around 9,320 people died in 2018 from this cancer in the USA. In addition, there were 91,270 new cases of melanoma diagnosed in the USA in 2018 [1].

Fortunately, if melanoma is detected in its early stages, then proper treatment can ensure a complete recovery. As a result, the survival rate for melanoma is very high. The early detection of this skin cancer is therefore critically important, and accurate equipment is necessary to ensure precise, early detection (even for highly trained experts).In the last decade, there has been a significant improvement in detecting melanoma skin cancer and these techniques have been successfully applied to ensure better treatment for different skin related problems. Generally, dermoscopic and clinical images are used for the analysis of skin cancer problems.

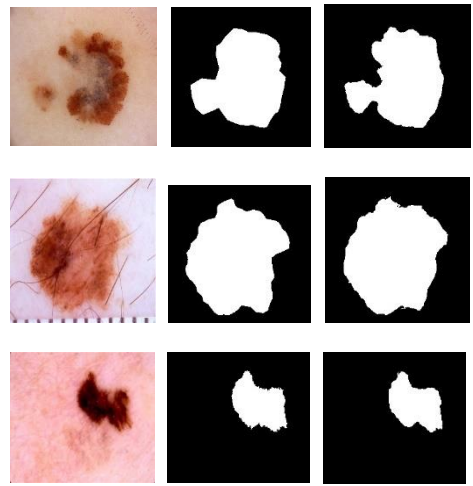

Figure 1. Medical image segmentation examples displaying input image segmentation on the left, ground truth (GT) for skin cancer lesion segmentation in the middle, and $\nabla^N$-Net outputs on the right.

A magnified and illuminated image of the skin cancer lesions is acquired with dermoscopy, which is a noninvasive skin imaging technique [2]. Dermoscopy imaging techniques enhance the visual effects of the regions of interest for obtaining very detailed, deeper levels of skin lesion regions through removing the surface reflection of the skin lesion [2]. It is reported that dermoscopy based skin cancer assessment systems show a higher accuracy compared to the naked eye [3]. Dermoscopy images are assessed by dermatologists and this tends to be very time consuming and error-prone process. This accuracy may drop drastically if clinicians have not been adequately trained [3].

Several studies show that the scarcity of dermatologists is another challenge towards ensuring better treatment or adequate levels of care for skin cancer patients. Therefore, an automatic skin cancer recognition system has a high demand to ensure accurate, faster and better treatment for skin cancer patients [2,3]. For clinical image analysis, different types of cameras, including mobile cameras, are used to acquire skin cancer lesion images. Varying orientations, illuminations, lighting conditions, and other artifacts make this problem difficult to analyze with automated approaches. Yang et al. [4] recently showed that this issue causes very low recognition accuracy in identifying diseases from various clinical images.

Due to several complexities, the automatic detection of melanoma skin cancer is a very challenging task. Firstly, there are large intraclass variances that can be observed in terms of color, textures, shape, size, contrast, and location. Additionally, there is a very high degree of similarity between melanoma and non-melanoma lesions, which makes this problem even harder. Secondly, at the very early stage of this cancer, the automatic skin lesion recognition task becomes more complex due to low contrast and obscuration between the affected areas on the skin and normal skin regions. Thirdly, several artifacts, including hairs, veins, ruler marks, and color calibration, may blur and occlude the skin cancer lesions, thus further reducing the recognition performance.

A recently developed DL based method shows significantly better performance for skin cancer classification, segmentation, and detection tasks. In 2017, an article was published in nature where DCNN based methods provide better testing accuracy compared to well-trained dermatologists (72.1% vs 66.0%), which truly exhibits the robustness of DCNN based methods [5]. In this paper, we propose and apply an improved DCNN model for skin cancer segmentation and classification. To accomplish our goal, the proposed model is evaluated with a set of experiments on dermoscopic images as shown in Figure 1. The contributions of this work can be summarized as follows:

- A new segmentation model, the NABLA-N network ($\nabla^N$-Net), is proposed and applied to a skin cancer segmentation task on the ISIC 2018 dataset.
- The impact of different fusion approaches in encoding and decoding units are investigated.
- The impact of transfer learning (TL) from ISIC-2017 to ISIC-2018 is evaluated for the segmentation task.
- An IRRCNN model is applied for skin cancer classification on the ISIC 2018 dermoscopic image dataset.

The paper is organized as follows: Section II discusses related work. The architectures of the proposed $\nabla^N$-Net model for the segmentation task and the IRRCNN model for classification task are presented in Section III. Section IV explains the datasets, experimental results, and analysis on results. The conclusion and future direction are discussed in Section V.

2. RELATED WORK

In the last few years, several methods have been proposed for skin cancer segmentation and classification problems. In 2016, at the International Skin Imaging Challenge (ISIC) different methods were proposed and submitted for skin cancer segmentation, feature extraction, and classification tasks. The final report published a comparative study and showed significantly higher segmentation and classification accuracies of 95.3% and 91.6% respectively. The classification results were based on recognizing only two types of cancer, benign and malignant [6]. In the following year (2017), a slightly bigger dataset compared to the 2016 version was released for segmentation, detection, and classification tasks (called ISIC 2017). Different methods were proposed and showed better recognition performance with different DCNNs and Support Vector Machine (L-SVM) methods. The Fully Convolutional Network (FCN) ensemble showed the highest segmentation performance in term of accuracy and dice coefficient which are 93.4% and 84.9% respectively [6,7].

In the same year, an automatic melanoma recognition system had been proposed which was applied to the ISBI 2016 dataset where a Fully Convolutional Residual Network (FCRN) was applied for accurate lesion segmentation [9]. After that, the lesion patches were cropped from whole input images and a Deep Residual Network (DRN) was used to classify melanoma and non-melanoma patches. In addition, experiments have been conducted with different DCNN methods including VGG-16, GoogleNet, FCRN-38, RCRN-50, and FCRN-101 [10]. The performance of VGG-16 and Inception-v3 were evaluated for skin lesion segmentation on the ISIC 2016 dataset, which showed around 61.3% and 69.3% testing accuracy as the highest performance in [10].

However, in 2015, an improved and new architecture, named "U-Net", was proposed specifically for medical image segmentation tasks. From then on, U-Net became very popular and was efficiently applied in different modalities of medical imaging and computational pathology. U-Net works with a smaller number of training samples while providing precise segmentation results [11]. In 2018, the Recurrent Residual Convolutional (RRN) U-Net was proposed, which is called R2U-Net [12]. This improved version of the U-Net model was tested on three different datasets for medical image segmentation tasks including a retinal blood vessel, skin cancer segmentation on ISIC 2017, and Lung segmentation datasets. The segmentation results were compared against SegNet, and Residual U-Net (ResU-Net). The quantitative and qualitative results showed significant improvement against the SegNet and the ResU-Net models for skin cancer segmentation tasks on the ISIC 2017 dataset [12]. Another network was proposed in 2018, which is called LadderNet. This improved architecture of U-Net model was essentially a chain of multiple U-Nets which contained multiple encoding and decoding units. This model can be viewed as a cascading of multiple FCNs and was tested for a retinal blood vessel segmentation task [13].

recently, in 2019, Fusion Net was proposed which consists of multiple U-Net models in parallel where the feature maps from the decoding unit of the first U-net is used with the inputs of the encoding unit of the second U-Net. Sigmoid gating layers were incorporated for attention modeling for the second U-Net. Experiments have been conducted on skin cancer segmentation on ISIC 2017 dataset and the results show better segmentation compared to R2U-Net [14]. In 2017, Nabla-Net was proposed, which consisted of encoding and decoding units that are based on FCNs which are applied for segmenting multiple sclerosis lesions and gliomas. This network essentially used the convolutional layers, with max-pooling for upsampling, and introduced tied unpooling layers [15]. In the encoding unit, the convolutional operations consisted of zero-paddings, 3x3 convolution, batch normalization followed by ReLU activation function. The zero paddings, 3x3 convolution, and Batch Normalization are used in the decoding unit. The feature maps from the second encoding layers are merged with concatenation to the 14$^{th}$ decoding layer. Finally, the sigmoid function is used to produce class confidence [15]. However, most of the published methods discussed above were evaluated on ISIC 2016 and 2017 datasets for segmentation tasks.

On the other hand, for the skin cancer classification task, several DL based methods were evaluated for the dermoscopic image on ISIC 2016 and 2017 datasets in [6,7]. The DL with established machine learning approaches makes the ensemble method. The lesions are extracted in the first step of classification, where the different classical machine learning methods (such as color histogram, edge histogram, sparse coding), and the DCNN based methods including ResNet, ImageNet, and U-Net shape were applied for feature extraction from input samples. Finally, non-linear SVM is applied for classification. These methods are tested on the ISIC 2016 dataset and shown that the method outperforms (76% vs 70.5%) when compared against 8 expert dermatologists on 100 subsets of test images [16]. The SVM and DCNN based methods were applied for skin cancer classification on a dataset which was collected from the web and showed promising recognition accuracy in [17]. The skin lesion analysis towards melanoma detection with deep learning methods was proposed in 2018. These DL based methods were evaluated on the ISIC 2017 dataset and the experimental results show around 0.718 dice coefficient for segmentation, 0.833 scores for feature extraction, and 0.823 scores for classification tasks [18]. The handcrafted feature-based approach and the DCNN model-based method are proposed for skin cancer recognition from clinical images for 198 different skin cancer problems [19].

In this paper, we propose a new model, named the NABLA-N network ($\nabla^N$-Net) which is applied for the skin cancer segmentation task. A pictorial representation of the model is shown in Figure 2. The model is explained below in section 3.1. This new proposed $\nabla^N$-Net model is significantly different from the existing NablaNet [15]. First, NablaNet is based on FCN whereas $\nabla^N$-Net is based on the R2U-Net model. Second, NablaNet uses only one latent space, however, the proposed $\nabla^N$-Net utilizes multiple latent spaces which essentially reuse the learned features. This concept helps significantly reduce the number of network parameters, which is inspired by the DenseNet model [20]. Third, in terms of feature fusion between encoding and decoding units, the NablaNet applies feature fusion between the second layer from the encoding unit to the 14$^{th}$ layer of the decoding unit. In contrast, the $\nabla^N$-Net applies different fusion among the features from the encoding and decoding units as well as decoding units itself. In addition, if we compare against the LadderNet and FusionNet. These use multiple U-Net models where the number of network parameters is increased significantly. On the other hand, the $\nabla^N$-Net model reuses the trained latent spaces from the encoding unit which helps to minimize the number of network parameters

*3.* DEEP CNN MODELS

3.1 NABLA-N($\nabla^N$-Net) Net for Skin Cancer Segmentation
The medical image segmentation models, including U-Net, ResU-Net, R2U-Net, LadderNet, and FusionNet models, consist of two units: encoding and decoding units. In the encoding unit, several layers of convolutional and subsampling operations are performed which produces different features represented in different stages of the encoding unit. The encoded features start to decode from the very bottleneck layers known as latent space. In this space, the number of feature maps is the highest and the dimension of the feature maps is the lowest. In the decoding unit, several transpose convolution operations are performed, and concatenation operations are performed between the features from encoding and decoding units.

According to the basic feature representation strategy of a convolutional network (classification model), during encoding of the input samples, the lower layers represent very low-level features (such as edges, corner, color, and so on). The deeper layers represent high-level features from the parts of the object to the entire shape of the object. Therefore, different features representation has significant importance in representing feature space to object space. However, for the segmentation task, we encode input samples and decode those from only considering the bottleneck layer in most of the cases. The decoding units are very crucial and sensitive to noise in producing accurate segmentation results. To enhance the ability to decode a unit to produce better and more accurate outputs, we propose the $\nabla^N$-Net model which is able to produce a better representation of features from the decoding units utilizing multiple feature spaces in the deeper layers of an encoding unit. However, the $\nabla^N$-Net consists of encoding and multiple decoding units. In the encoding unit, the inputs are encoded according to the other models, but the multiple decoding units decode the encoded features from different latent spaces. Since the architecture is similar to the inverse delta symbol, we have named this model as NABLA (which is a triangular symbol like an inverted Greek Delta) network. The N is defined based on the number of feature spaces that are used for implementing NABLA. Figure 2 shows the entire $\nabla^3$-Net architecture with three decoding units used.

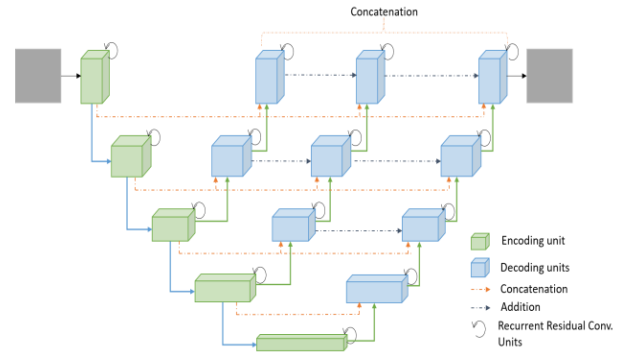

Figure 2. $\nabla^3$-Net CNN model semantic bio-medical image segmentation consists of encoding and decoding units. Three encoded feature spaces are used.

Based on the different feature fusion methods of the proposed model, three different architectures are evaluated in this implementation, named $\nabla^N$-Net$_A$, $\nabla^N$-Net$_B$, and $\nabla^N$-Net$_{AB}$. In $\nabla^N$-Net$_A$ we apply concatenation operations between the features from encoding and decoding units, indicated with the orange line in Figure 2. At the actual output space, the concatenation operations are performed on decoded features from multiple units followed by 1×1 convolution to produce the final outputs. In this model, no feature fusion operations are performed in the decoding units themselves. For $\nabla^N$-Net$_B$, we have applied concatenation between the feature maps from encoding and decoding units. In addition, we have applied feature fusion with added operations between decoding units, indicated with a blue line in Figure 2. The architecture with the combination of the outputs from both feature fusion methods is named $\nabla^N$-Net$_{AB}$.

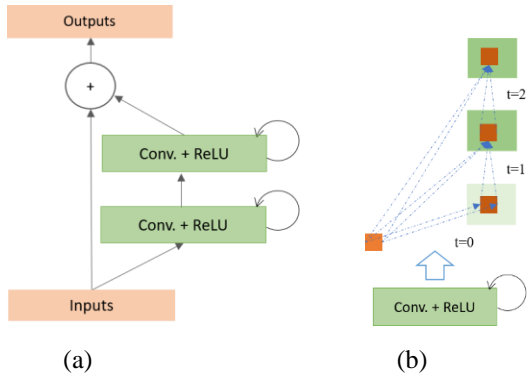

(a)          (b)

Figure 3. Different variants of the convolutional and recurrent convolutional units including (a) the Recurrent Residual Convolutional Unit (RRCU).

For convolutional operations, we perform recurrent residual convolutional operations in both encoding and decoding units, shown in Figure 3. This method has already proved significant advantages against U-Net and ResU-Net for a different application of medical image segmentation tasks [12]. There are several advantages of RRUs. First, the recurrent residual network helps to train deeper and bigger models. Second, it ensures better feature representation that leads to better performance for classification, segmentation, and detection tasks. Third, it ensures better performance with the same number of network parameters [12]. The robustness of the proposed model and the advantages of different feature fusion between encoding and decoding units are evaluated with a set of experiments for segmentation tasks on the ISIC 2018 dermoscopic image dataset.

3.2 IRRCNN for Skin Cancer Classification

In this work, we apply the Inception Recurrent Residual Convolutional Neural Network (RRCNN) model for skin cancer recognition from dermoscopic images. The IRRCNN network model shows superior testing performance compared to equivalent inception and inception of residual networks for object recognition tasks in several computer vision problems. The details of the IRRCNN model is described in [21]. To our knowledge, this is the first time the IRRCNN model is applied for skin cancer classification. The IRRCNN architecture consists of several inceptions recurrent residual units (IRRUs) which are shown in Figure 3 (a). The recurrent operation is performed with respect to different time steps, shown in Figure 3(b).

3.3 Network architectures

For the segmentation model, we use the following:
1→16(3)→32(3)→64(3)→128(3)→256(3)→512(3)→256(3)→128 (3)→64(3)→32(3)→16(3)→1 where the number outside of the parenthesis indicates the number of feature maps and the number inside the parenthesis represents the filter size of that respective layer. The total number of network parameters of $\nabla^N$-Net is around 18.78 Million (M).

The IRRCNN model consists of 3 inception recurrent residual units followed by sub-sampling layers. At the end of this model, we used a Global Average Pooling (GAP) layer followed by a softmax layer. The GAP layer helps to reduce the number of network parameters significantly compared to a fully connected layer. The IRRCNN models utilize only 11.2M network parameters in this implementation.

Table 1. The testing accuracy R2U-Net and $\nabla^N$-Net and comparison against existing methods. Bold text indicates the best results.

| Method | Precision | Recall | Accuracy | F1-score | IoU | DI |
|---|---|---|---|---|---|---|
| R2U-Net | 0.9629 | 0.9556 | 0.9556 | 0.9553 | 0.8719 | 0.8841 |
| $\nabla^2$-Net$_A$ | 0.9631 | 0.9545 | 0.9545 | 0.9551 | 0.8688 | 0.8777 |
| $\nabla^2$-Net$_B$ | 0.9642 | 0.9570 | 0.9570 | 0.9572 | 0.8751 | 0.8877 |
| $\nabla^2$-Net$_{AB}$ | 0.9659 | 0.9582 | 0.9581 | 0.9582 | 0.8803 | 0.8927 |
| $\nabla^2$-Net$_{AB}$ + TL | **0.9707** | **0.9636** | **0.9636** | **0.9644** | **0.8883** | **0.8960** |
| $\nabla^2$-Net$_{AB}$ + TL + Data Aug. | 0.9668 | 0.9603 | 0.9603 | 0.9603 | 0.8821 | 0.8929 |

## 4. RESULTS AND DISCUSSION

To demonstrate the performance of the segmentation (including $\nabla^N$-Net$_A$, $\nabla^N$-Net$_B$, and, $\nabla^N$-Net$_{AB}$) and classification models, we have evaluated them for skin cancer segmentation and classification tasks. For this implementation, a TensorFlow DL framework is used on a single GPU machine with 56GB of RAM and an NVIDIA GEFORCE GTX-1080 Ti GPU with 12GB of memory.

### 1. Evaluation metrics

For quantitative analysis of the experimental results, several performance metrics were considered, including precision, recall, accuracy (AC), F1-score, Intersection over Union (IoU), and Dice Coefficient (DI). To do this we also use the variables True Positive (TP), True Negative (TN), False Positive (FP), and False Negative (FN). The precision and recall are expressed as:

$$\text{Precision} = \frac{TP}{TP+FP} \quad (1)$$

$$\text{Recall} = \frac{TP}{TP+FN} \quad (2)$$

The overall accuracy is calculated using Eq. (3),

$$AC = \frac{TP+TN}{TP+TN+FP+FN} \quad (3)$$

In addition, we have also conducted an experiment to determine that the IoU that is represented using Eq. (4). Here GT refers to the ground truth and SR refers to the segmentation result.

$$IoU = \frac{|GT \cap SR|}{|GT|+|SR|} \quad (4)$$

The F1-Score is calculated according to the following equation:

$$F1-score = 2 \times \frac{\text{precision} \times \text{recall}}{\text{precision}+\text{recall}} \quad (5)$$

Furthermore, the Dice Coefficient (DI) is calculated using the following Eq. (6).

$$DI = \frac{2.TP}{2.TP+FN+FP} \quad (6)$$

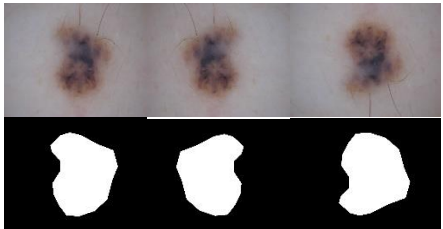

Figure 4. Data augmentation for segmentation task: original image on the left, horizontal and vertical flipping samples are shown on the middle and right respectively.

### 2. Databases

In both segmentation and classification, we have used dermoscopic images from ISIC 2018.

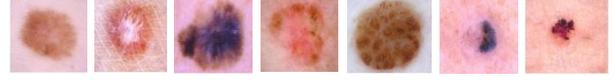

Figure 5. Example images for seven different skin cancer types.

Segmentation dataset: The total number of images is 2,594 with corresponding masks, where 80 percent (2100) of samples are used to train and validate the model and the remaining 20 percent (494) of samples are used for testing. The original size of the images is approximately from 767×576 pixels to 6748×4499 pixels. Due to hardware limitations, we have resized to 256×256, therefore the input samples lose a significant amount of information. The target pixels are set to a value of either 255 or 0, denoting pixels inside or outside the target lesion respectively.

Classification dataset: In this experiment, we have used the ISIC 2018 dataset for skin cancer. The total number of samples is 10,015 and the size of the images are 650×450 pixels. The images are downsampled to 192×192 pixels where 70 percent of the samples (7,010) are used for training and validation and the remaining 30 percent of the samples (3005) are used for testing. The previous version of the ISIC datasets had only two different classes (benign and malignant) whereas the ISIC 2018 dataset contains seven different classes of skin cancers. The seven different classes of skin cancers are Nevus, Dermatofibroma, Melanoma, Pigmented Bowen's, Pigmented Benign Keratoses, Basal Cell Carcinoma, and Vascular which are shown in Figure 5 from the left to the right in order.

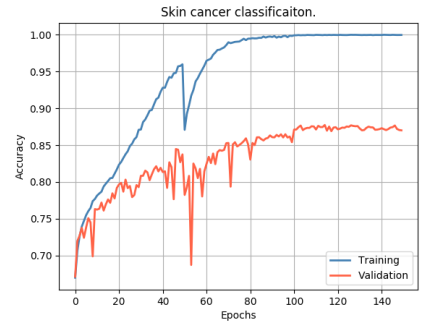

Figure 6. Training and validation accuracy for skin cancer classification with IRRCNN method.

### 3. Training method

The end-to-end $\nabla^N$-Net models are trained for segmentation tasks with Adam optimizer with a learning rate of ($3 \times e^{-4}$) and binary cross entropy loss. The model is trained for 250 epochs with a batch size of 8. For the classification problem, the IRRCNN model is trained for 150 epochs in total with batch size 8 and an initial learning rate of 0.01. The learning rate is decreased after every 50 epochs by a factor of 10. We have used a momentum of 0.9. The Stochastic Gradient Descent

(SGD) optimization method and categorical cross entropy loss are used in this implementation. The training and validation accuracy of the IRRCNN model are shown in Figure 6.

## 5. RESULTS AND DISCUSSION

We have evaluated R2U-Net and three $\nabla^N$-Net models for skin cancer segmentation. In addition, we have investigated the performance of the proposed model with the Transfer Learning (TL) method from ISIC 2017 to ISIC 2018 and with data augmentation. We have applied horizontal and vertical flipping for data augmentation, as shown in Figure 4. Data augmentation increases the number of training samples by a factor of three. The quantitative results and comparison are shown in Table 1. The $\nabla^2$-Net$_{AB}$ model provides better testing accuracy compared to R2U-Net, $\nabla^2$-Net$_A$, and $\nabla^2$-Net$_B$. The $\nabla^2$-Net$_{AB}$ model with TL shows the highest testing accuracy in term of global accuracy and DC which is around 1.55% and 0.33% improvement compared to the highest testing accuracy without TL. The model is also trained and tested for augmented data; however, we did not observe any significant improvement of data augmentation during testing.

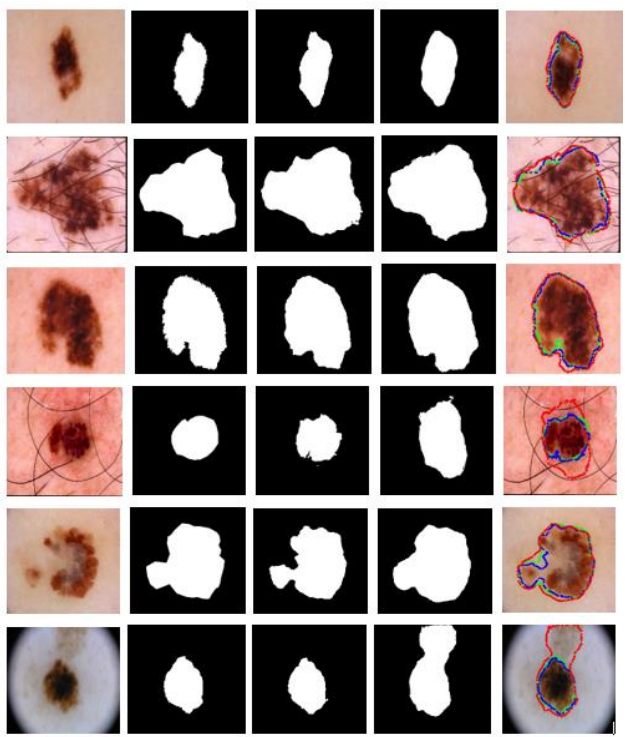

Inputs    GT    $\nabla^N$-Net$_{AB}$    R2U-Net    Contours

Figure 7. The first column shows the inputs, the second column shows the Ground Truth (GT), the third column shows the outputs from $\nabla^N$-Net$_{AB}$, fourth columns show the outputs for R2U-Net, and the fifth column shows the results with respective contours for GT, R2U-Net, and $\nabla^N$-Net models where green color represents the GT, blue color for $\nabla^2$-Net$_{AB}$, and red color shows contour for R2U-Net model.

For qualitative analysis, the testing results of R2U-Net and $\nabla^2$-Net$_{AB}$ are shown in Figure 7. From this figure, it is observed that both R2U-Net and $\nabla^N$-Net$_{AB}$ provide good segmentation results. However, in some difficult cases, the $\nabla^2$-Net$_{AB}$ model provides better segmentation lesions with proper contours compared to R2U-Net. Figure 7 shows the experimental outputs with R2U-Net and $\nabla^N$-Net$_{AB}$ networks where the first column shows the inputs, the second column shows the ground truth (GT), the third column shows outputs from $\nabla^N$-Net network, the fourth column represents the outputs from R2U-Net, and the fifth column represents contours of GT, $\nabla^N$-Net$_{AB}$ and R2U-Net model with green, blue, and red colors respectively. From the outputs, it can be seen that the $\nabla^2$-Net$_{AB}$ network provides better segmentation results with appropriate contour compared to R2U-Net. Thus, it is shown that the feature fusion between encoding and decoding units along with the feature's fusion in the decoding unit itself helps to improve testing performance. Additionally, if we observe the contour of the segmentation outputs, it can be seen that $\nabla^N$-Net$_{AB}$ provides superior performance which is very close to the GT (in some cases, it is better). Furthermore, as the $\nabla^2$-Net$_{AB}$ model provides the best quantitative and qualitative results, we investigated the impact of a number of feature spaces (latent space) that is used in decoding from the encoding unit. The different version $\nabla^N$-Net$_{AB}$ models are evaluated where N = 1,2,3, and 4. We have achieved 0.8842, 0.8960, 0.8941, and 0.8949 DI scores for $\nabla^1$-Net$_{AB}$, $\nabla^2$-Net$_{AB}$, $\nabla^3$-Net$_{AB}$, and $\nabla^4$-Net$_{AB}$ models respectively. The $\nabla^2$-Net$_{AB}$ model provides the highest result in term of accuracy and DI. This indicates that even if we perform fusion from different feature spaces toward the low level features of encoding units, no significant impact is observed on the testing accuracy.

Table 2. The precision, recall, and F1-score for skin cancer classification with data augmentation.

| Classes | Precision | Recall | F1-score | Support |
|---|---|---|---|---|
| 0 | 0.76 | 0.51 | 0.61 | 326 |
| 1 | 0.91 | 0.95 | 0.93 | 2008 |
| 2 | 0.85 | 0.82 | 0.83 | 160 |
| 3 | 0.82 | 0.71 | 0.76 | 103 |
| 4 | 0.72 | 0.80 | 0.76 | 328 |
| 5 | 0.84 | 0.75 | 0.79 | 37 |
| 6 | 0.95 | 0.93 | 0.94 | 43 |
| Average | 0.87 | 0.87 | 0.86 | 3005 |

For the skin cancer classification task, the performance is evaluated both with data augmentation and without data augmentation. We have applied a horizontal and vertical flipping method for data augmentation. The results show that data augmentation helps to improve performance for classification tasks significantly. The plot for training and validation accuracy for augmented samples is shown in Figure 6. We have tested on 3,005 samples for seven different skin cancer types. The model shows 81.12% testing accuracy for a dataset without data augmentation, whereas we have achieved 87.09% testing accuracy when the model is trained on augmented samples. Therefore, data augmentation helps to improve the performance of around 6%. The average of precision, recall, and F1-score and the number of supporting samples during the testing phase are shown in Table 2. The weighted average precision, recall, and F1-scores are 0.87, 0.87, and 0.86 respectively.

The computational time for segmentation models takes around 55 to 64 seconds processing time per epoch during training and only average 6 seconds for testing of 494 samples. On the other hand, the classification model takes 430 seconds for training per epoch and around 38 seconds for testing of 2,005 samples.

6. CONCLUSION

In this paper, we have proposed an improved U-Net which is named as NABLA-N ($\nabla^N$-Net) and the model is evaluated for skin cancer segmentation tasks. Three different $\nabla^N$-Net models are investigated with different feature fusion between encoding and decoding units which are evaluated on the ISIC2018 dataset. The quantitative and qualitative results demonstrate better performance with the $\nabla^2$-Net$_{AB}$ model compared to the R2U-Net model. In addition, the Inception Recurrent Residual Convolutional Neural Network (IRRCNN) model is applied for skin cancer classification on the ISIC 2018 database. The IRRCNN shows 87.09% testing accuracy on augmented samples. Further investigations include a more experimental evaluation with different feature fusion in decoding units and testing on more complex 3D samples, including Neuroimaging, and CT, etc.